\documentclass[runningheads]{llncs}
\usepackage{threeparttable}
\usepackage{enumitem}
\usepackage{graphicx}
\usepackage{booktabs}
\usepackage{array}
\usepackage[T1]{fontenc}
\usepackage[labelfont=bf, justification=justified, singlelinecheck=true]{caption}
\usepackage{subcaption}
\usepackage{amssymb}
\usepackage{xcolor}
\usepackage{makecell}
\usepackage{amsmath}
\usepackage{float}
\usepackage{cite}
\usepackage[colorlinks=true, allcolors=blue]{hyperref}

\usepackage{marvosym}

\definecolor{ForestGreen}{RGB}{34,139,34}

\newcommand{\SoM}{\textit{Set-of-Mark }}
\newcommand{\VLLMs}{Vision Large Language Models }
\begin{document}

\title{Leveraging Gaze and Set-of-Mark in VLLMs for Human-Object Interaction Anticipation from Egocentric Videos}
\titlerunning{Leveraging Gaze and Set-of-Mark in VLLMs for HOI Anticipation}
\author{Daniele Materia\inst{1}\textsuperscript{(\Letter)} \and Francesco Ragusa\inst{1,2} \and Giovanni Maria Farinella\inst{1,2}}
%\author{Daniele Materia\inst{1}\textsuperscript{(\Letter)}\orcidID{0009-0008-3815-7333} \and Francesco Ragusa\inst{1,2}\orcidID{0000-0002-6368-1910} \and Giovanni Maria Farinella\inst{1,2}\orcidID{0000-0002-6034-0432}}
\authorrunning{D. Materia et al.}

\institute{Department of Mathematics and Computer Science, University of Catania, Italy \and Next Vision s.r.l. -- Spinoff of the University of Catania, Italy \\ \email{daniele.materia@studium.unict.it}}

\maketitle

\begin{abstract}
The ability to anticipate human-object interactions is highly desirable in an intelligent assistive system in order to guide users during daily life activities and understand their short and long-term goals. Creating systems with such capabilities requires to approach several complex challenges. This work addresses the problem of human-object interaction anticipation in Egocentric Vision using \VLLMs (VLLMs). We tackle key limitations in existing approaches by improving visual grounding capabilities through \textit{Set-of-Mark} prompting and understanding user intent via the trajectory formed by the user's most recent gaze fixations. To effectively capture the temporal dynamics immediately preceding the interaction, we further introduce a novel \textit{inverse exponential sampling} strategy for input video frames. 
Experiments conducted on the egocentric dataset HD-EPIC demonstrate that our method surpasses state-of-the-art approaches for the considered task, showing its model-agnostic nature.
We release the code at the following \href{https://github.com/fpv-iplab/leveraging_gaze_som_vllms_human_obj_anticipation}{GitHub repository}.


\end{abstract}

\section{Introduction}
Understanding human behavior from an egocentric point of view enables the development of systems that can support individuals in their daily activities across various domains \cite{plizzari2024outlookfutureegocentricvision}, including kitchens \cite{damen2018scalingegocentricvisionepickitchens,li2020eyebeholdergazeactions}, sports and hobbies \cite{grauman2022ego4dworld3000hours}, as well as industrial workplaces \cite{ragusa2023enigma51finegrainedunderstandinghumanobject,ragusa_MECCANO_2023}.
The ability to accurately anticipate human-object interactions is particularly valuable for such systems, as it enables them to assist users and predict their intentions \cite{Furnari_2017_NAO,Furnari_2021_RU-LSTM}. For example, in the kitchen domain, the system should alert the user if they are about to touch a hot pot without using pot holders \cite{damen2018scalingegocentricvisionepickitchens,ragusa2024stillfastendtoendapproachshortterm,perrett2025hdepichighlydetailedegocentricvideo}. Similarly, in industrial workplaces, forecasting potentially hazardous interactions can help prevent accidents by alerting workers and thereby improving safety~\cite{grauman2022ego4dworld3000hours,grauman2025egoexo4d,ragusa2023enigma51finegrainedunderstandinghumanobject}.

Toward this direction, previous works have investigated different aspects of anticipation, focusing on predicting hand trajectories \cite{jia2022generative,Ma2025NovelDM},
actions \cite{Furnari_2021_RU-LSTM,ragusa_MECCANO_2023,Roy_Wacv24,Rodin2025Egocentric}, user localization \cite{park_ego_loc,future_loc_makansi}, and objects and their interactions \cite{Furnari_2017_NAO,perrett2025hdepichighlydetailedegocentricvideo,thakur2024anticipatingactiveobjectsegocentric,ragusa2024stillfastendtoendapproachshortterm,grauman2022ego4dworld3000hours}.

With the recent advent of Vision Large Language Models (VLLMs), the anticipation task has been increasingly investigated using these models due to their ability to address a wide range of challenges. In particular, recent studies have focused on predicting future actions \cite{palm_forecasting_eccv24,wang2025actionllm}, motion tracking \cite{Hong_2025_CVPR}, and \mbox{human-object} interactions \cite{perrett2025hdepichighlydetailedegocentricvideo}.

\begin{figure}[t!]
    \centering
    \includegraphics[width=1\linewidth]{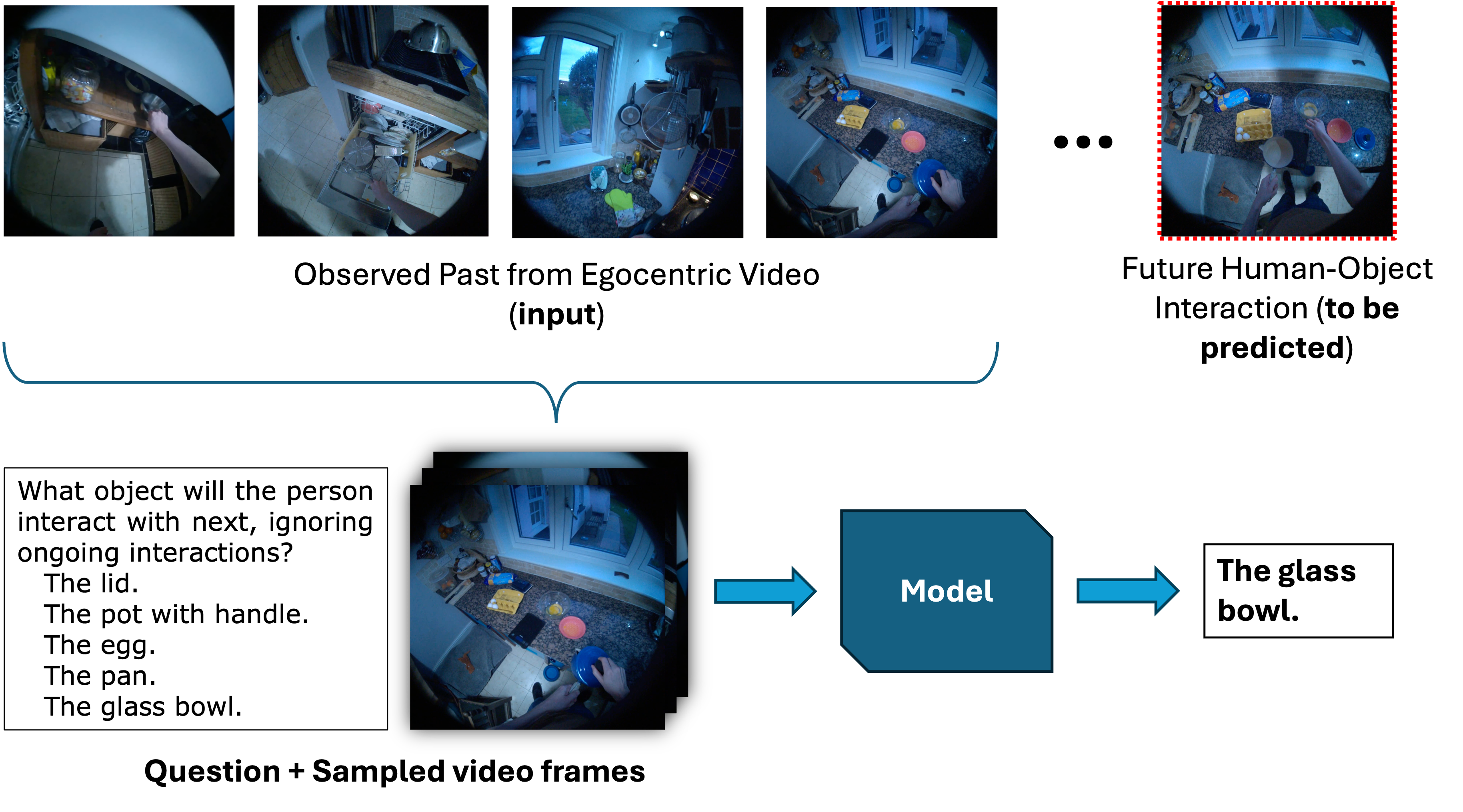}
    \caption{Conceptual scheme of the VQA Interaction Anticipation task.}
    \label{fig:task_repr}
\end{figure}

In this paper, we address the problem of human-object interaction anticipation from egocentric videos leveraging the capabilities of VLLMs. We formulate this task as Visual Question Answering (VQA), adopting the protocol established by the recent \textit{HD-EPIC Gaze Interaction Anticipation} benchmark \cite{perrett2025hdepichighlydetailedegocentricvideo}. Specifically, the model is queried to distinguish the interaction's target object from a set of candidates, included in the input prompt, based on the visual context (see Figure~\ref{fig:task_repr}). To this end, we propose a novel strategy to enhance the visual grounding capabilities of VLLMs by incorporating spatial scene information through \SoM (SoM) prompting \cite{yang2023setofmarkpromptingunleashesextraordinary} and modeling human intent via user gaze trajectories.
Furthermore, we introduce an effective video sampling strategy for the visual input, aimed at optimizing the information required to anticipate human-object interactions while discarding irrelevant data. The proposed approach is model-agnostic, meaning it can be applied to any VLLM.

Experiments on the HD-EPIC dataset \cite{perrett2025hdepichighlydetailedegocentricvideo} show that the proposed method outperforms state-of-the-art approaches, highlighting that the adopted models significantly benefit from our proposed strategy to anticipate human-object interactions. Additionally, we performed an ablation study to assess the impact of each component of the proposed approach on the overall performance.

The contributions of this work are as follows: 1) we propose a novel approach that incorporates spatial scene information and models user intention through gaze trajectories; 2) through extensive experiments on the HD-EPIC dataset, our method outperforms state-of-the-art approaches in anticipating human-object interactions; 3) we release the source code of our approach at the following \href{https://github.com/fpv-iplab/leveraging_gaze_som_vllms_human_obj_anticipation}{GitHub repository} to encourage future research in this area.

\section{Related Work}
\label{sec:related_work}

\subsection{Egocentric Datasets}
\label{subsec:datasets}
Research in Egocentric Vision has been significantly accelerated in the last years by the release of several datasets. Initial efforts were made by the authors of the ADL (Activities of Daily Living) dataset \cite{pirsiavash2012detecting}, which featured 20 videos of subjects performing unscripted daily routine activities like making coffee or washing dishes in their own homes, providing a rich source of complex human-object interactions. The authors of \cite{damen2018scalingegocentricvisionepickitchens} released the first large-scale egocentric dataset: EPIC-KITCHENS. With 55 hours of video and 32 participants involved, it includes dense annotations for 39.6K action segments (spanning 125 verbs and 352 noun classes) and 454K object bounding boxes. Focusing exclusively on unscripted kitchen activities, this pioneering work was a turning point for the field, despite considering solely visual data. The dataset was later expanded to 100 hours of recordings in \cite{Damen2022RESCALING}, further consolidating its status as a standard benchmark for the community.

EGTEA Gaze+ \cite{li2020eyebeholdergazeactions} was among the first datasets to incorporate user gaze information, providing it alongside egocentric video to support action recognition. The MECCANO dataset \cite{ragusa2021meccanodatasetunderstandinghumanobject} focused on the industrial-like domain, specifically capturing recordings of subjects assembling a toy motorbike model. It comprises 20 videos from 20 subjects and includes 8,857 temporal annotations, involving 61 action classes, 12 verbs and 20 object classes. Its extension, MECCANO Multimodal \cite{ragusa_MECCANO_2023} is one of the earliest examples of multimodal egocentric dataset, providing synchronized recordings of RGB, depth, and gaze data. Additionally, the authors emphasize the utility of the gaze signal in the proposed \textit{Action Recognition} and \textit{Active Object Detection} benchmarks, where they explicitly compare RGB-only models against RGB-Gaze models, demonstrating that integrating gaze helps the model focus on relevant visual features.

The ENIGMA-51 dataset \cite{ragusa2023enigma51finegrainedunderstandinghumanobject} was designed to bridge the gap to real-world industrial environments, focusing on the fine-grained understanding of human-object interactions during the repair of electrical boards. It comprises 51 videos featuring dense, \mbox{fine-grained} annotations including 14,036 interactions and 275K object bounding boxes, making it a challenging benchmark for human-object interaction detection in industrial environments.
Later, the scale of egocentric research was redefined by Ego4D \cite{grauman2022ego4dworld3000hours}, a massive collaborative effort comprising 3,670 hours of video, collected by 923 participants across 9 countries; covering a vast range of scenarios beyond just the kitchen or industry, including household chores, workplace activities, and leisure. The authors introduced a benchmark composed of five tasks: \textit{Episodic Memory}, \textit{Hands and Objects}, \textit{Audio-Visual Diarization}, \textit{Social Interactions}, and \textit{Forecasting}. This effort was subsequently expanded by Ego-Exo4D \cite{grauman2025egoexo4d}, which focuses on skilled human activities, such as sports, music, and dance, captured simultaneously from time-synchronized egocentric and exocentric perspectives. It provides rich annotations including expert commentary and 3D body pose to support benchmarks like proficiency estimation.

A significant step forward was marked by the release of HD-EPIC \cite{perrett2025hdepichighlydetailedegocentricvideo}, a highly detailed dataset designed to push the boundaries of fine-grained video understanding. It features 41 hours of unscripted video of kitchen activities. The dataset stands out for its annotation density and multimodality, featuring 59K fine-grained actions, 51K audio events, 20K object movements, 37K object masks lifted to 3D and the users' gaze data. A key contribution is its challenging VQA benchmark of 26K questions, which tests capabilities ranging from recipe understanding to interaction anticipation. Finally, the landscape has been further enriched by the release of Ego-EXTRA \cite{ragusa2025egoextravideolanguageegocentricdataset}. Designed to evaluate VLLMs in expert-trainee scenarios, it features 50 hours of unscripted procedural tasks where trainees are guided by remote experts. Unique to this dataset is the \textit{Wizard of Oz} collection protocol that allows to capture high-quality, bidirectional natural language dialogue aligned with multimodal signals, including the gaze of both the trainee and the expert.

Among all of these, we chose HD-EPIC as our benchmark dataset due to its multimodality, dense annotations and because of the presence of the specific \textit{Gaze Interaction Anticipation} benchmark.

\subsection{Human-Object Interaction Anticipation}
\label{subsec:object_anticipation_literature}
Research in this area has been explored in various forms. A pioneering work on human-object interaction anticipation was introduced by the authors of \cite{Furnari_2017_NAO}, who first defined the problem as \textit{Next-Active Object} anticipation. Their approach leverages the analysis of object motion trajectories within the egocentric view to distinguish between \textit{active} objects (those likely to be interacted with) and \textit{passive} background elements. Subsequently, the authors of \cite{zhanggazeanticipationadversarialnetworks} employed a Generative Adversarial Network (GAN) to predict future gaze locations, utilizing this estimated visual attention as a spatial prior for action anticipation. Similarly, the authors of \cite{nagarajan2019groundedhumanobjectinteractionhotspots} focused on anticipating object affordances by predicting interaction hotspots; these cues were subsequently leveraged by \cite{liu2020forecastinghumanobjectinteractionjoint}, who proposed a joint modeling approach where the motor attention (future hand trajectory) and visual attention (interaction hotspots) are predicted simultaneously, allowing the two tasks to mutually reinforce the final action prediction.

Later, the object anticipation task was standardized by the authors of \cite{grauman2022ego4dworld3000hours} and named \textit{Short-Term Object-Interaction Anticipation}. It consisted of predicting which object would be interacted with next (noun), how the interaction would take place (verb), and when it would begin (time-to-contact). Building on this formalization of the task, the authors of \cite{pasca2024summarizepastpredictfuture} employed a multimodal transformer architecture that leverages pretrained vision-language and captioning models to extract the action context from past frames, enabling object-interaction anticipation. Another contribution was made by the authors of \cite{ragusa2024stillfastendtoendapproachshortterm} proposing a CNN-based model able to simultaneously process a single image and a video to detect and localize the next-active object, predict the interaction verb, and estimate the time-to-contact.

Lastly, the authors of \cite{perrett2025hdepichighlydetailedegocentricvideo} introduced the \textit{Gaze Interaction Anticipation} benchmark, as a subset of the broader \textit{HD-EPIC VQA benchmark}. Reflecting the steady rise of multimodal large language models, they evaluated the performance of several state-of-the-art VLLMs by formulating their benchmark as a visual question answering (VQA) task, where models select the correct answer from a list of candidates; therefore, the task requires the model to predict the noun of the target object, distinguishing the true one from semantic distractors based on the visual context.

In this work, we propose a novel approach that enhances the capabilities of VLLMs by incorporating spatial information, gaze trajectories, and a new video sampling strategy. The proposed method is model-agnostic and achieves state-of-the-art performance on the \textit{HD-EPIC Gaze Interaction Anticipation} benchmark.

\section{Proposed Approach}
\label{sec:approach}

\begin{figure}[t!]
    \centering
    \includegraphics[width=1.0\linewidth]{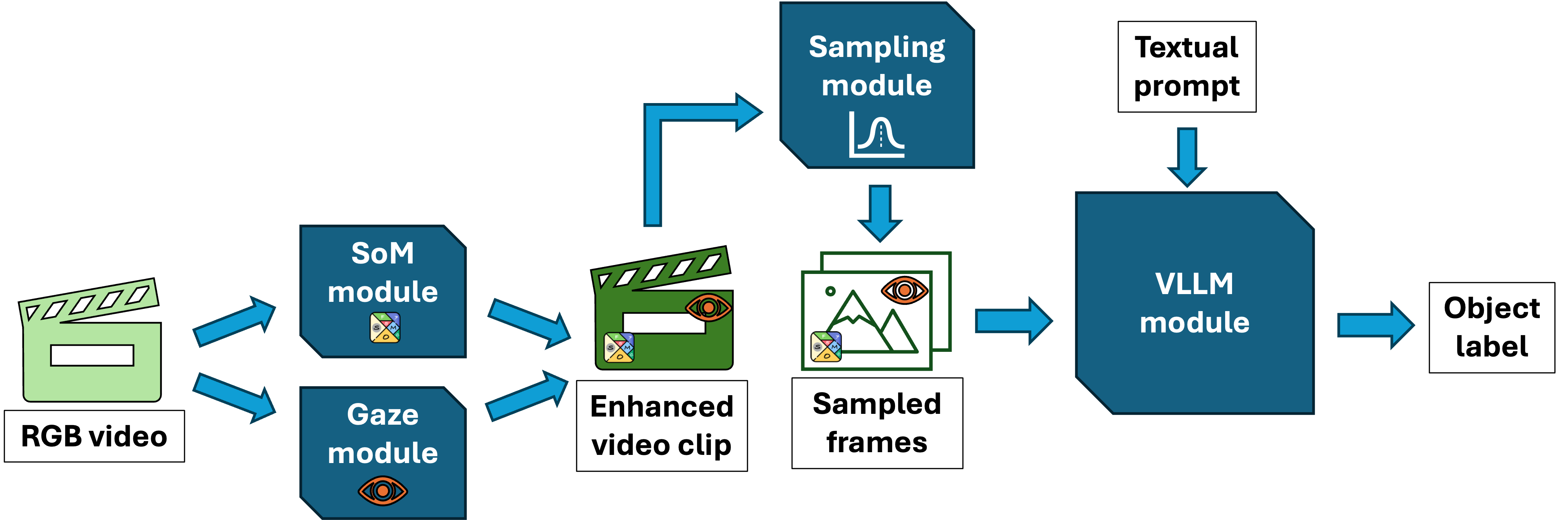}
    \caption{The proposed architecture for the human-object interaction anticipation task. It is composed of 4 main modules: 1) \SoM Module, 2) Gaze Module, 3) Sampling Module, and 4) VLLM Module.}
    \label{fig:pipeline}
\end{figure}

The proposed approach consists of four key modules for predicting the next human-object interaction: 1) the \SoM Module, which enhances visual grounding and scene understanding; 2) the Gaze Module, which gives signal about user intention through gaze trajectories; 3) the Sampling Module, which optimizes the visual input by retaining relevant information and discarding redundant data; 4) the VLLM Module, which performs the prediction of the next human-object interaction.
Figure~\ref{fig:pipeline} shows the overall proposed architecture.

\newpage
\subsection{Set-of-Mark Module}
\SoM~\cite{yang2023setofmarkpromptingunleashesextraordinary} is a visual prompting method that partitions an input image into semantic regions at varying levels of granularity. These regions are then overlaid with semantic masks and, optionally, alphanumeric marks or bounding boxes.
Several VLLMs have demonstrated a robust capability to interpret these visual markers, either as an emergent property or through targeted visual instruction tuning ~\cite{li2024llavaonevisioneasyvisualtask,yang2023setofmarkpromptingunleashesextraordinary,wu2024visualpromptingmultimodallarge}, enabling them to effectively leverage SoM and therefore enhance their spatial grounding capabilities.

\begin{figure}[t]
    \centering
    \begin{subfigure}[b]{0.248\textwidth}
        \centering
        \includegraphics[width=\textwidth]{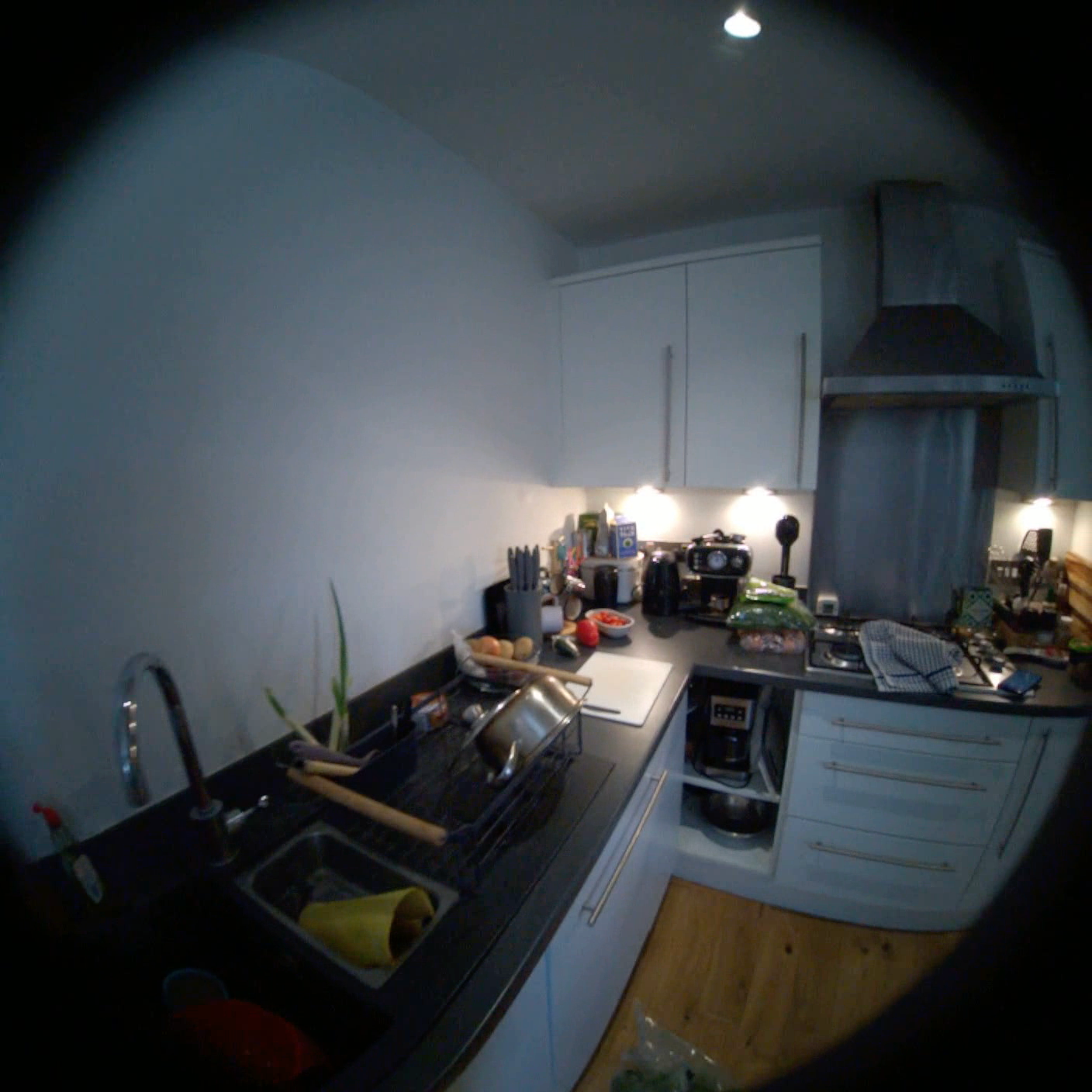}
        \caption{}
    \end{subfigure}%
    \hfill
    \begin{subfigure}[b]{0.248\textwidth}
        \centering
        \includegraphics[width=\textwidth]{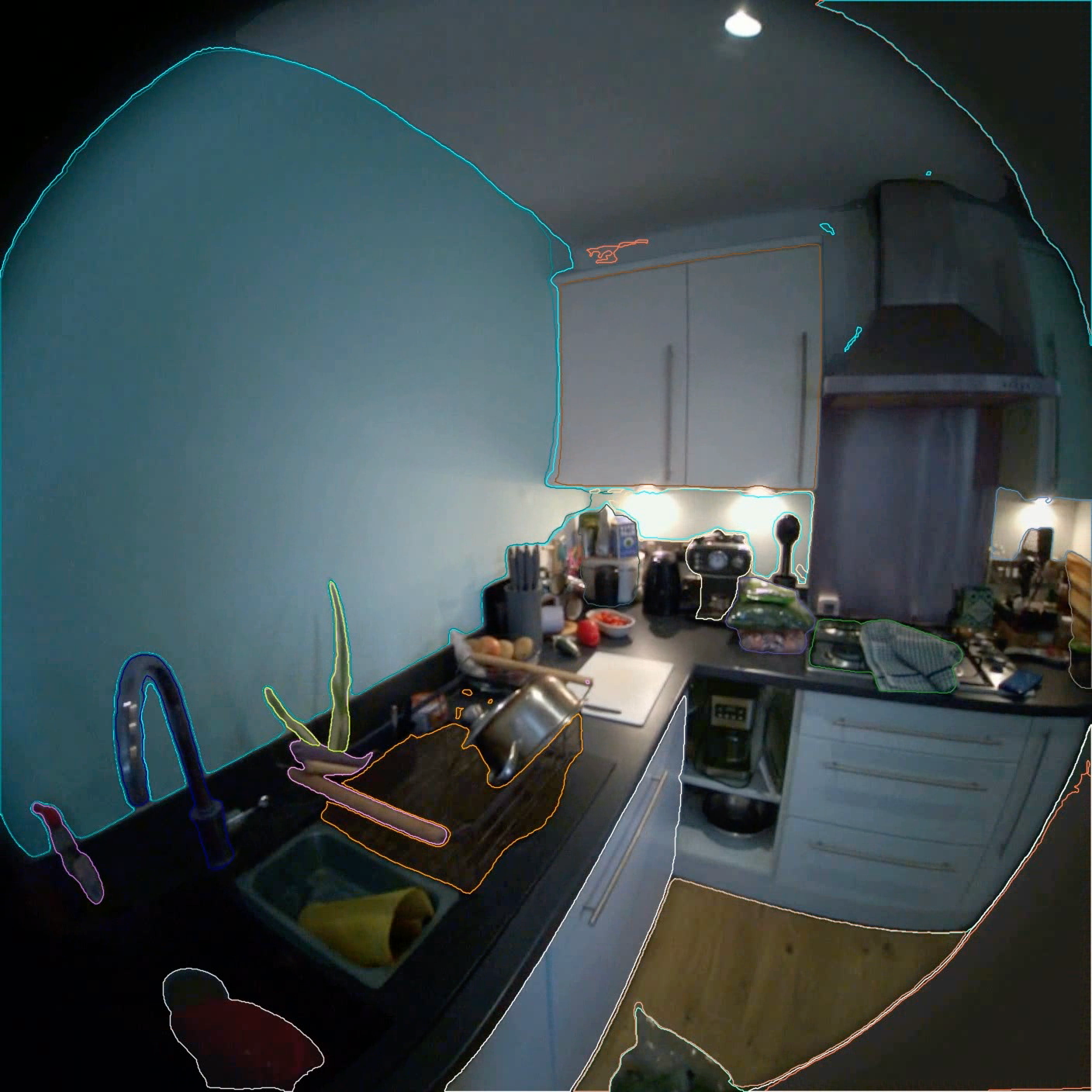}
        \caption{}
    \end{subfigure}%
    \hfill
    \begin{subfigure}[b]{0.248\textwidth}
        \centering
        \includegraphics[width=\textwidth]{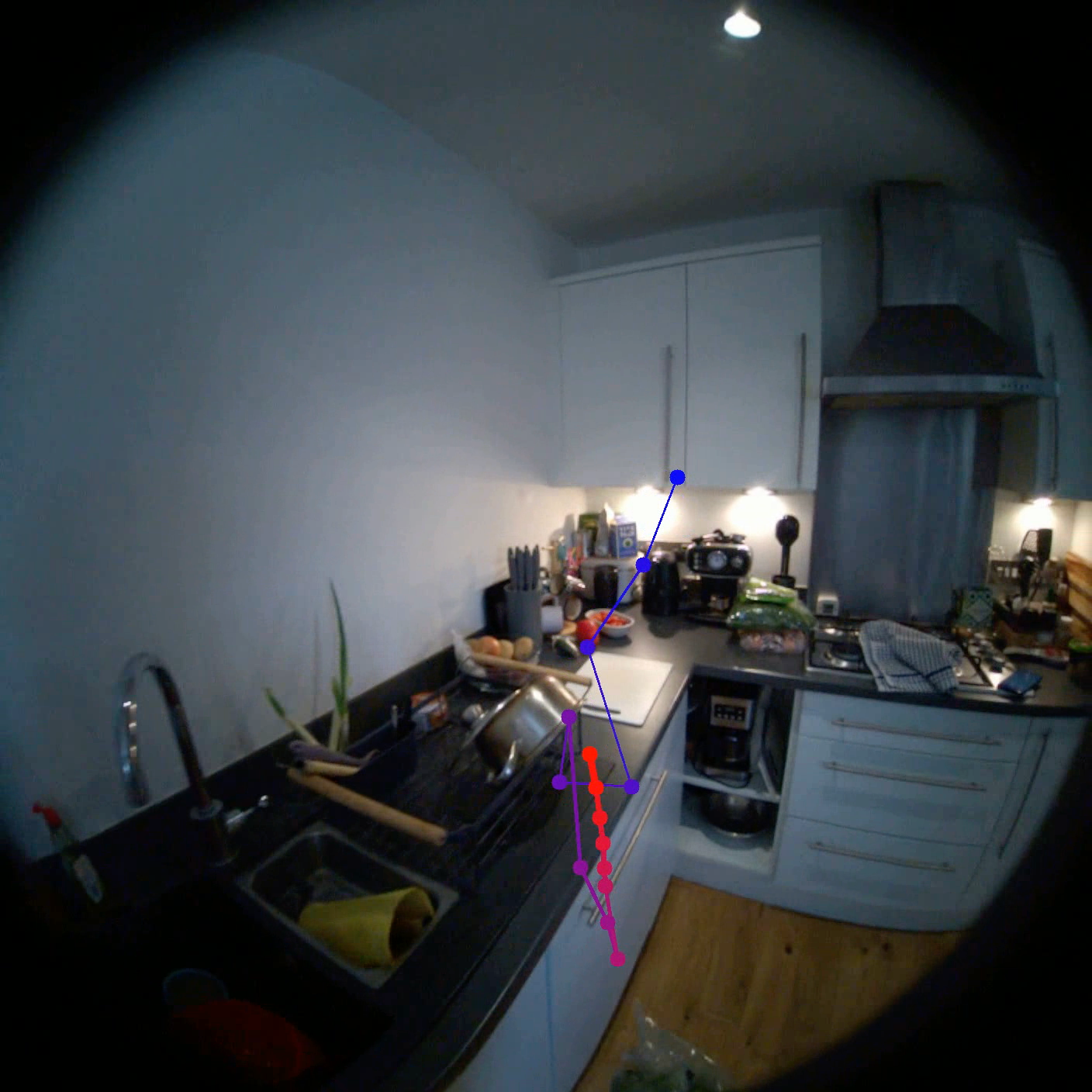}
        \caption{}
    \end{subfigure}%
    \hfill
    \begin{subfigure}[b]{0.248\textwidth}
        \centering
        \includegraphics[width=\textwidth]{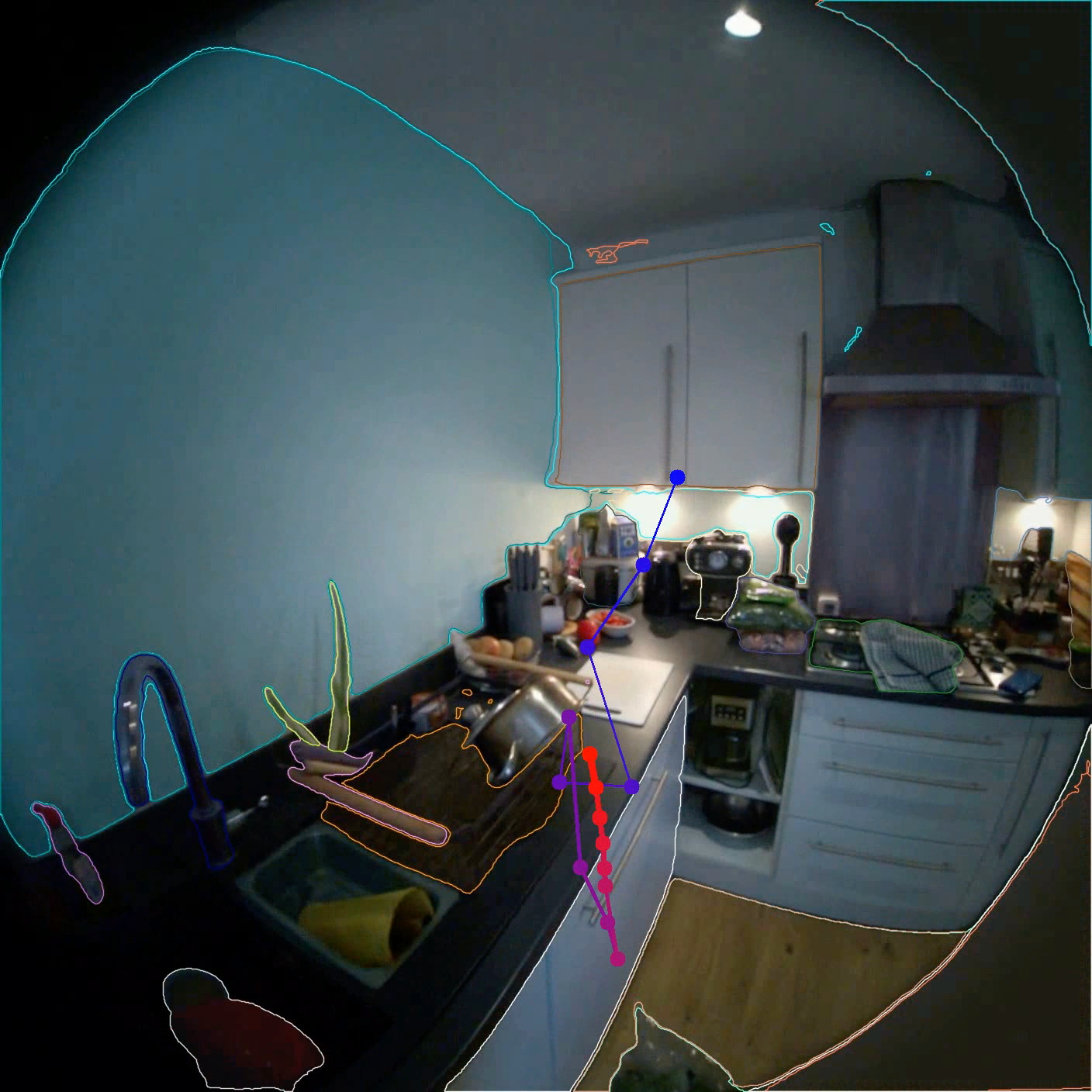}
        \caption{}
    \end{subfigure}
    
    \caption{The figure illustrates how the input frame is processed by the SoM and Gaze modules. The input RGB frame (a) is first processed by the SoM Module (b) and the Gaze Module (c). Their outputs are then fused (d) to obtain a visual representation that incorporates both spatial and intention-related information.}
    \label{fig:strategies_example}
\end{figure}

The proposed SoM Module operates exclusively on the final frame of the input video, generating semantic masks via Semantic-SAM ($\alpha =0.05$ transparency) to enrich the scene's visual representation. Notably, we omit alphanumeric tags, as segmentation masks alone are sufficient for visual grounding given that the model predicts the target object's name from a candidate list contained in the input prompt.
Figure~\ref{fig:strategies_example}-b shows an example of the output of the SoM Module.

\subsection{Gaze Module}
Inspired by the role of eye gaze in understanding human intentions during activities from visual input \cite{roleofvisiondailyliving,Yarbus_1967}, and considering its exploitation in egocentric vision for identifying attended objects \cite{Mazzamuto2023learning}, discovering object usage \cite{Damen2015YouDoIU}, and predicting human-object interactions \cite{zhou2024interattn}, we leverage the gaze signal as a prior to anticipate human-object interactions.

Specifically, for each frame $F_t$, we consider a sliding window to project the sequence of the most recent $W=15$ ground-truth 2D gaze fixations available up to time $t$. This trajectory is rendered using connected circles that gradually change color from red (most recent fixation) to blue (least recent fixation), providing a visual cue that helps the model infer the user’s attention and intent. Figure~\ref{fig:strategies_example}-c shows the output of the Gaze Module.

\subsection{Sampling Module}
Transformer architectures exhibit a time and space complexity of $O(N^2)$ with respect to the input sequence length $N$ \cite{vaswani2017attentionneed}. For VLLMs, which map visual data into a backbone LLM's embedding space as long sequences of tokens, this makes processing dense video inputs computationally prohibitive, strictly limiting the number of input frames. To ensure this selected subset of frames is as informative as possible, we prioritize the temporal window immediately preceding the interaction through \textit{inverse exponential sampling}. This strategy biases input frame selection toward the end of the video clip, enabling the model to focus on the most relevant temporal context for human-object interaction anticipation.
\begin{figure}[t!]
    \centering
\includegraphics[width=1\linewidth]{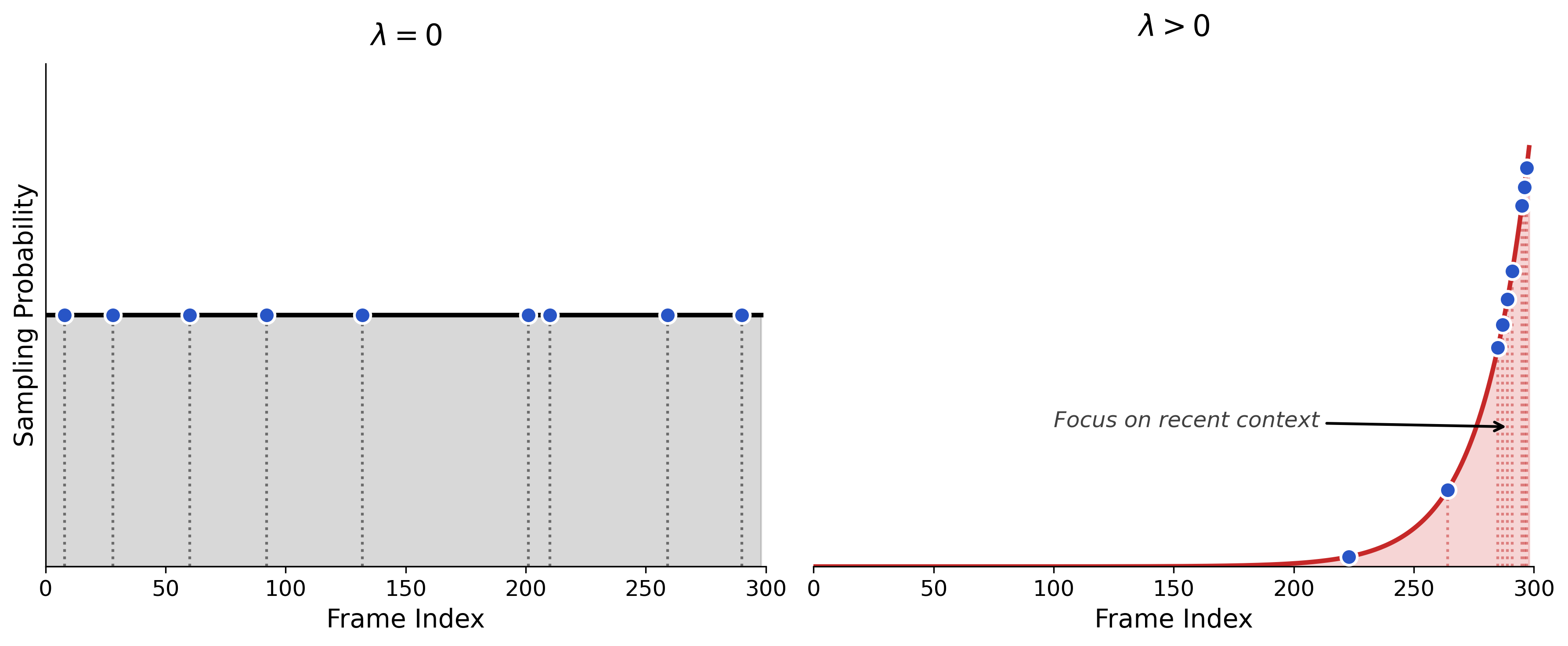}
    \caption{Visualization of our \textit{inverse exponential sampling} strategy with $n=10$. Setting $\lambda=0$ (left) results in a uniform distribution, while $\lambda > 0$ (right) concentrates the sampled frames (the blue dots) on the instants immediately preceding the interaction. The blue dots represent the $n-1$ probabilistically sampled frames; the final frame, which is always part of the input, is omitted for clarity.}
    \label{fig:inv_exp_sampling_example}
\end{figure}

Formally, given a video sequence of length $N_v$ and a sample size $n$, we treat the final frame ($i = N_v - 1$) as a mandatory input to guarantee the inclusion of the most recent visual information. To select the remaining $n-1$ frames (indices $0 \leq i \leq N_v - 2$), we define the selection probability $p_i$ for the $i$-th frame as:
\begin{equation}
p_i = \frac{e^{-\lambda (N_v - 2 - i)}}{\sum_{j=0}^{N_v - 2} e^{-\lambda (N_v - 2 - j)}}, \quad 0 \leq \lambda \leq 1
\label{eq:inv-exp-sampling}
\end{equation}
where the hyperparameter $\lambda$ controls the strength of the temporal bias towards the end of the sequence. As $\lambda$ increases, the probability distribution assigns higher sampling probability to the final indices of the sequence, which are closer to the human-object interaction we want to anticipate. When $\lambda = 0$, the distribution collapses to a uniform one, where for each frame $i$:
\begin{equation}
    p_i = \frac{1}{N_v-1}
\end{equation}

Figure~\ref{fig:inv_exp_sampling_example} visualizes the proposed sampling strategy and the effect of $\lambda$ on the distribution. We analyze the impact of the sample size $n$ in Subsection~\ref{subsec:ablation}.

\subsection{VLLM Module}
The VLLM Module constitutes the final step of our pipeline. Specifically, the model receives the $n$ sampled video frames, enhanced through SoM and Gaze Modules, alongside a textual prompt comprising the question, candidate answers, and instructions regarding the visual cues (see Figure~\ref{fig:pipeline}). It then outputs the predicted target object. Since our approach operates directly on the input frames, this design allows for the seamless integration of any state-of-the-art VLLM.

\section{Experimental Settings and Results}
\label{sec:experiments}

\subsection{Dataset}
For our experiments, we adopt the \textit{Gaze Interaction Anticipation} subset of the HD-EPIC VQA benchmark. It comprises $1,000$ multiple-choice questions paired with their corresponding 10-second video clips. Each video is carefully trimmed to end $0.3$ seconds after the user's gaze primes the object intended for the subsequent interaction.
Each question includes a correct answer, representing the object with which the person will interact, and four negative answers randomly sampled from all other objects that have been moved within the 2 minute video segment \cite{perrett2025hdepichighlydetailedegocentricvideo} (see Figure~\ref{fig:task_repr}-bottom for an example). To better align the visual input, obtained with our approach, with the VLLM's reasoning capabilities, we include specific additional instructions in the base prompt. 
In particular, we explicitly instruct the model to base its prediction on the final frame (where the segmentation masks are applied) while using the preceding frames as temporal context, as well as, we provide a detailed description of the gaze trajectory, explaining that the red circles and connected paths represent the user's fixation history, and direct the model to prioritize objects along this trajectory. Figure \ref{fig:qualitative_519} reports an example of the adopted prompt. As evaluation measures, we followed the HD-EPIC \textit{Gaze Interaction Anticipation} benchmark and adopted accuracy to assess model performance.

\subsection{VLLMs Selection}
Since our approach is model-agnostic, we adopted two different VLLMs in our experiments.
We have benchmarked \textbf{LLaVA-OneVision} \cite{li2024llavaonevisioneasyvisualtask} as our primary VLLM due to its strong video understanding capabilities, essential for effective egocentric human-object interaction anticipation. Specifically, its 7B variant ensures high performance with manageable computational resources.

Furthermore, to ensure a fair comparison
with the HD-EPIC baseline (Gemini 1.5 Pro) and to demonstrate that the observed performance improvements are independent of the VLLM used, we also adopted  \textbf{Gemini 2.0 Flash} \cite{google2025aiapi,google_gemini_2_0_flash}. This model was chosen as the successor to the now-deprecated Gemini 1.5 Pro, in accordance with Google's official migration guidelines \cite{geminimodelversions}.

\begin{table}[t]
    \centering
    \caption{Results for the Interaction Anticipation task. The best results are reported in \textbf{bold}, whereas the second best results are \underline{underlined}.}
    \begin{tabular}{lc} 
    \toprule 
        Approach & Accuracy (\%) \\
    \midrule
        K-Net \cite{taluzzi2025pixelsgraphsusingscene} & 15.7 \\
        T-CoT \cite{hd_epic_challenge_second} & 15.8 \\
        LLaVA-OneVision 7B \cite{li2024llavaonevisioneasyvisualtask} & 20.4\\
        Gemini 1.5 Pro (HD-EPIC baseline) \cite{perrett2025hdepichighlydetailedegocentricvideo} & 21.0 \\
        Qwen2.5VL-Ricoh \cite{hd_epic_challenge_third} & 22.0\\
    \bottomrule
        Ours (with LLaVA-OneVision 7B) & \underline{27.2} \\
        Ours (with Gemini 2.0 Flash) & \textbf{27.5} \\
    \end{tabular}
    \label{tab:comparison_results}
\end{table}

\subsection{Results}
We evaluated the performance of our proposed approach against current state-of-the-art models on the \textit{HD-EPIC Gaze Interaction Anticipation} benchmark. Specifically, we compared our model with the HD-EPIC baseline, as well as with the top 3 approaches that participated in the HD-EPIC VQA challenge (K-Net\cite{taluzzi2025pixelsgraphsusingscene}, T-CoT\cite{hd_epic_challenge_second}, and Qwen2.5VL-Ricoh\cite{hd_epic_challenge_third}). We also report the performance of LLaVA-OneVision in its standard setting \cite{llava_onevision_transformers} (i.e., without visual cues and with $n=8$ uniformly sampled video frames) to serve as a direct baseline. Table~\ref{tab:comparison_results} reports the obtained results.
The proposed approach consistently outperforms SOTA methods, achieving \textbf{27.2\%} and \textbf{27.5\%} accuracy when integrated with LLaVA-OneVision and Gemini 2.0 Flash, respectively (last two rows). These results set a new state-of-the-art for the \textit{HD-EPIC Gaze Interaction Anticipation} benchmark, surpassing the previous best performance obtained by Qwen2.5VL-Ricoh of more than 5\%.
It is worth noting that the improvements are not related to a specific VLLM, observing gains of \textbf{+6.8\%} for LLaVA-OneVision and \textbf{+6.5\%} for Gemini compared to their respective standard counterparts (third and fourth row). This demonstrates the model-agnostic nature of our architecture and its ability to enhance VLLMs' interaction anticipation capabilities.

\subsection{Qualitative Results}
\begin{figure}[t!]
    \centering
    \begin{minipage}[c]{0.35\textwidth}
        \small
        \textbf{Prompt:} Focus on the last frame to make your prediction and use the rest of the video to infer the context. (\textit{if SoM is used})
        
        What object will the person interact with next, ignoring ongoing interactions?

        \begin{list}{}{
                \setlength{\leftmargin}{0.5em}
                \setlength{\labelwidth}{0pt}
                \setlength{\labelsep}{0pt}
                \setlength{\itemsep}{0.1em}
                \setlength{\topsep}{0.05cm}
                \setlength{\partopsep}{0pt}
                \setlength{\parsep}{0pt}
            }
            \item The lid.
            \item The pot with handle.
            \item The egg.
            \item The pan.
            \item The glass bowl.
        \end{list}

        Follow the user's gaze trajectory closely: the red circles indicate where the user has most recently looked, and the connected path shows the sequence of gaze points across the most recent frames. The objects that have just been fixated are very likely to include the one the user will interact with next. Use this visual cue to make your prediction.\newline(\textit{if Gaze is used})
    \end{minipage}%
    \hfill
    \begin{minipage}[c]{0.63\textwidth}
        \centering
        \begin{subfigure}[b]{0.48\linewidth}
            \centering
            \includegraphics[width=\textwidth]{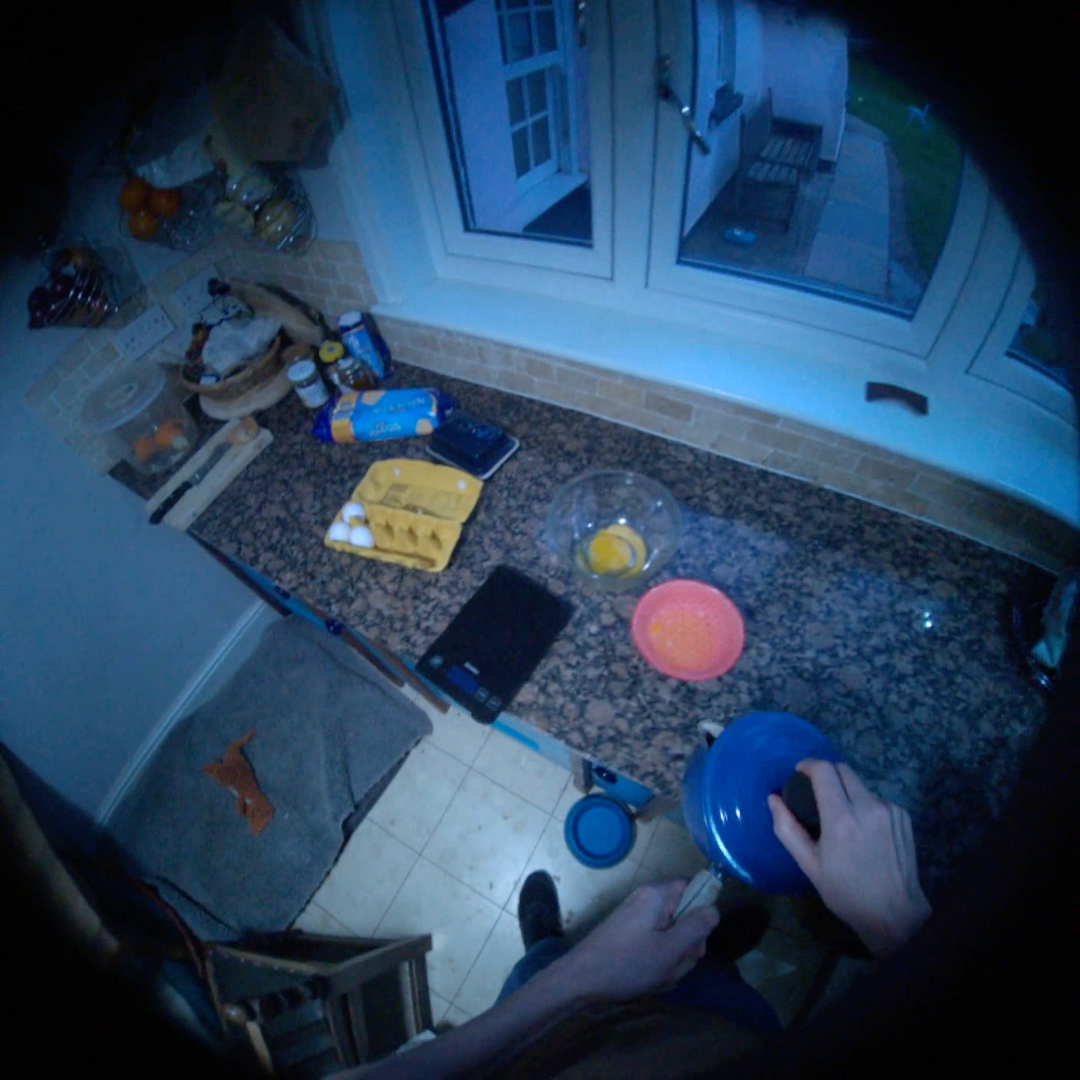}
            \colorbox{red!10}{\parbox{0.94\linewidth}{\centering\scriptsize\textcolor{red}{\textbf{The lid.}}}}
            \caption{}
        \end{subfigure}\hfill
        \begin{subfigure}[b]{0.48\linewidth}
            \centering
            \includegraphics[width=\textwidth]{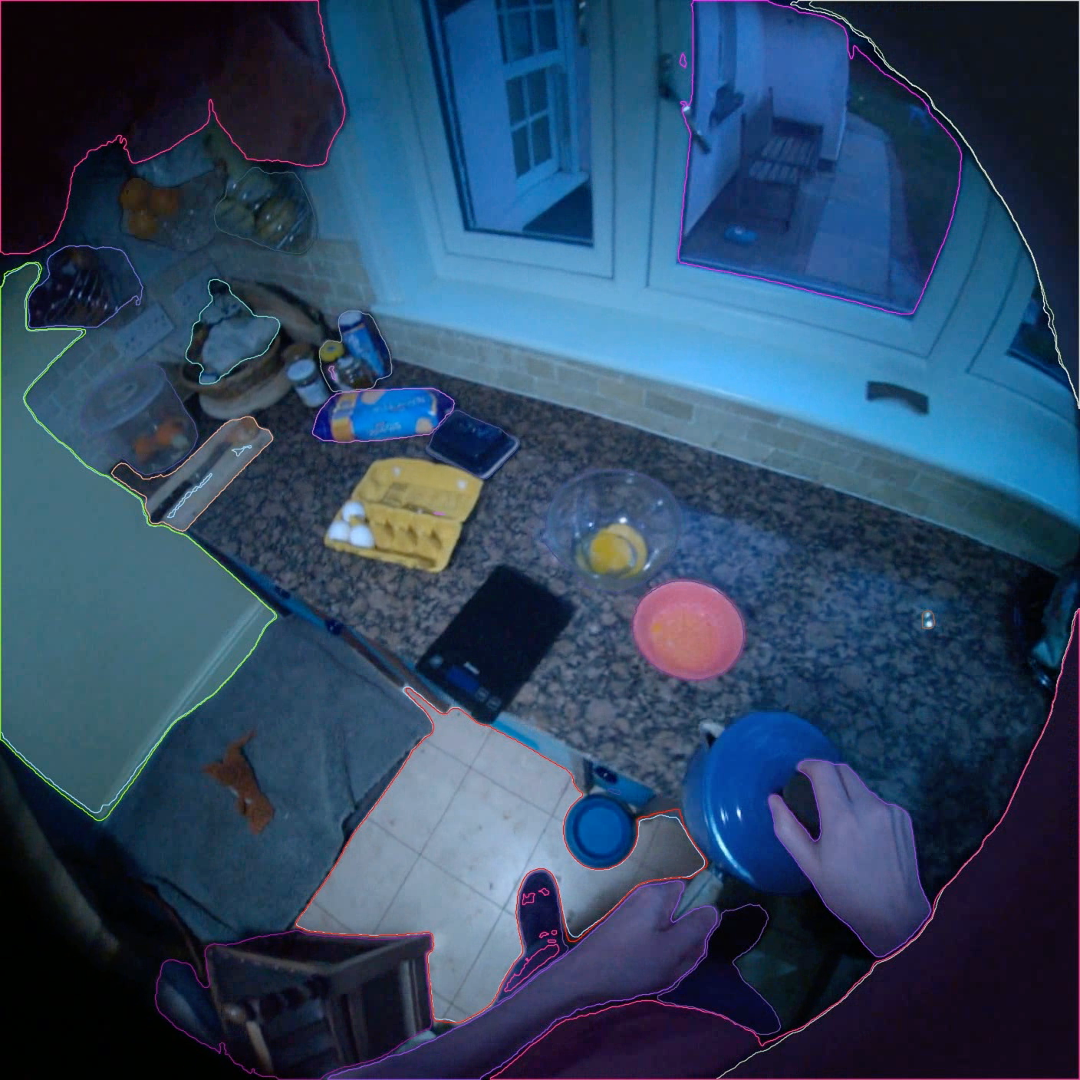}
            \colorbox{red!10}{\parbox{0.94\linewidth}{\centering\scriptsize\textcolor{red}{\textbf{The lid.}}}}
            \caption{}
        \end{subfigure}
        
        \begin{subfigure}[b]{0.48\linewidth}
            \centering
            \includegraphics[width=\textwidth]{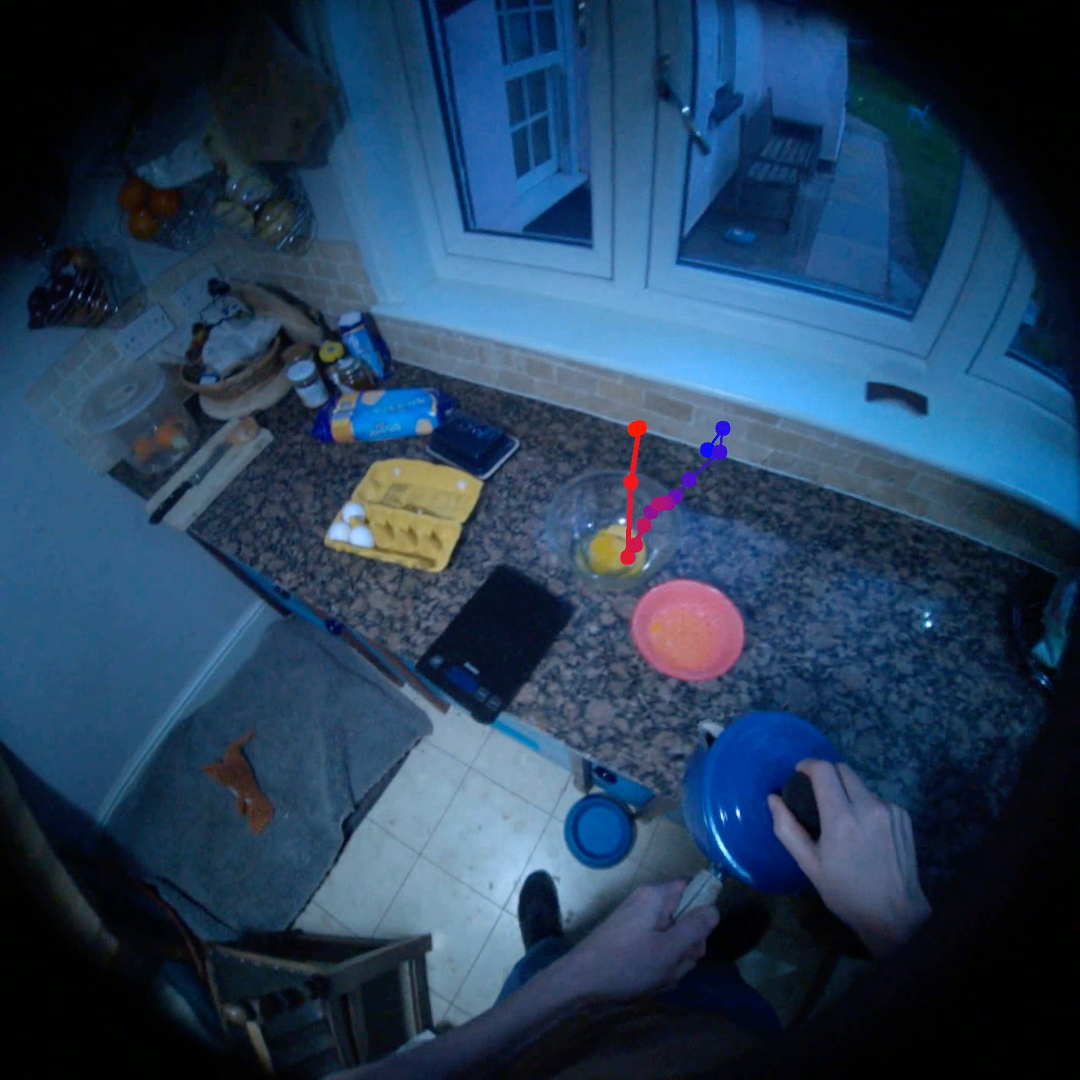}
            \colorbox{green!10}{\parbox{0.94\linewidth}{\centering\scriptsize\textcolor{green!60!black}{\textbf{The glass bowl.}}}}
            \caption{}
        \end{subfigure}\hfill
        \begin{subfigure}[b]{0.48\linewidth}
            \centering
            \includegraphics[width=\textwidth]{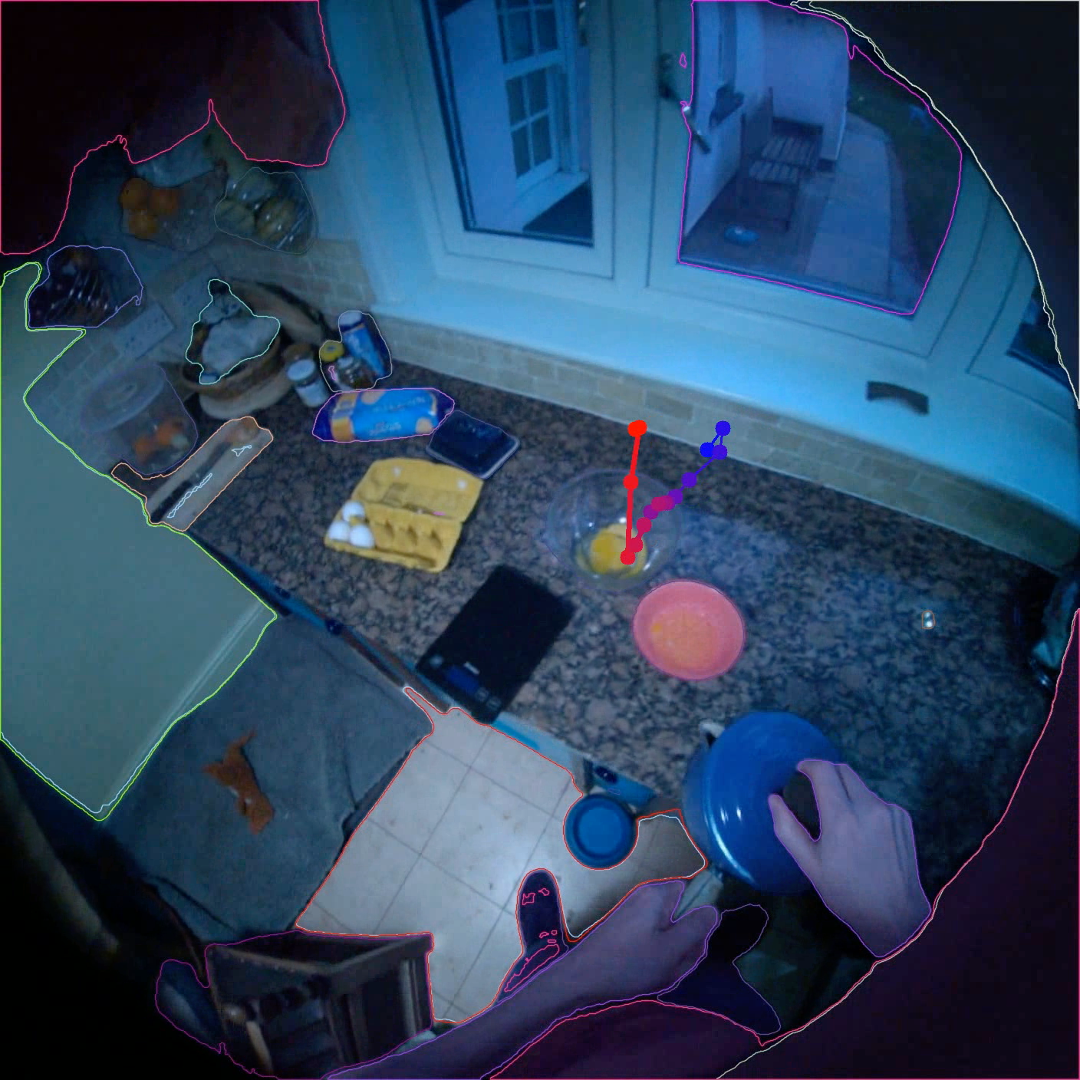}
            \colorbox{green!10}{\parbox{0.94\linewidth}{\centering\scriptsize\textcolor{green!60!black}{\textbf{The glass bowl.}}}}
            \caption{}
        \end{subfigure}
    \end{minipage}
    
    \caption{Qualitative example showing the positive impact of incorporating gaze information. (a) Classic VLLM, (b) proposed approach considering VLLM and SoM modules, (c) proposed approach considering VLLM and Gaze modules, (d) proposed approach considering VLLM, SoM and Gaze modules.}
    \label{fig:qualitative_519}
\end{figure}

\begin{figure}[ht!]
    \centering
    \begin{minipage}[c]{0.35\textwidth}
        \small
        \textbf{Prompt:} Focus on the last frame to make your prediction and use the rest of the video to infer the context. (\textit{if SoM is used})
        
        What object will the person interact with next, ignoring ongoing interactions? 

        \begin{list}{}{
                \setlength{\leftmargin}{0.5em}
                \setlength{\labelwidth}{0pt}
                \setlength{\labelsep}{0pt}
                \setlength{\itemsep}{0.1em}
                \setlength{\topsep}{0.05cm}
                \setlength{\partopsep}{0pt}
                \setlength{\parsep}{0pt}
            }
            \item The orange wiping cloth.
            \item The green capped condiment jar.
            \item The marker.
            \item The white serving bowl.
            \item The cap of pen.
        \end{list}

        Follow the user's gaze trajectory closely: the red circles indicate where the user has most recently looked, and the connected path shows the sequence of gaze points across the most recent frames. The objects that have just been fixated are very likely to include the one the user will interact with next. Use this visual cue to make your prediction.\newline(\textit{if Gaze is used})
    \end{minipage}%
    \hfill
    \begin{minipage}[c]{0.63\textwidth}
        \centering
        \begin{subfigure}[t]{0.48\linewidth}
            \centering
            \includegraphics[width=\textwidth]{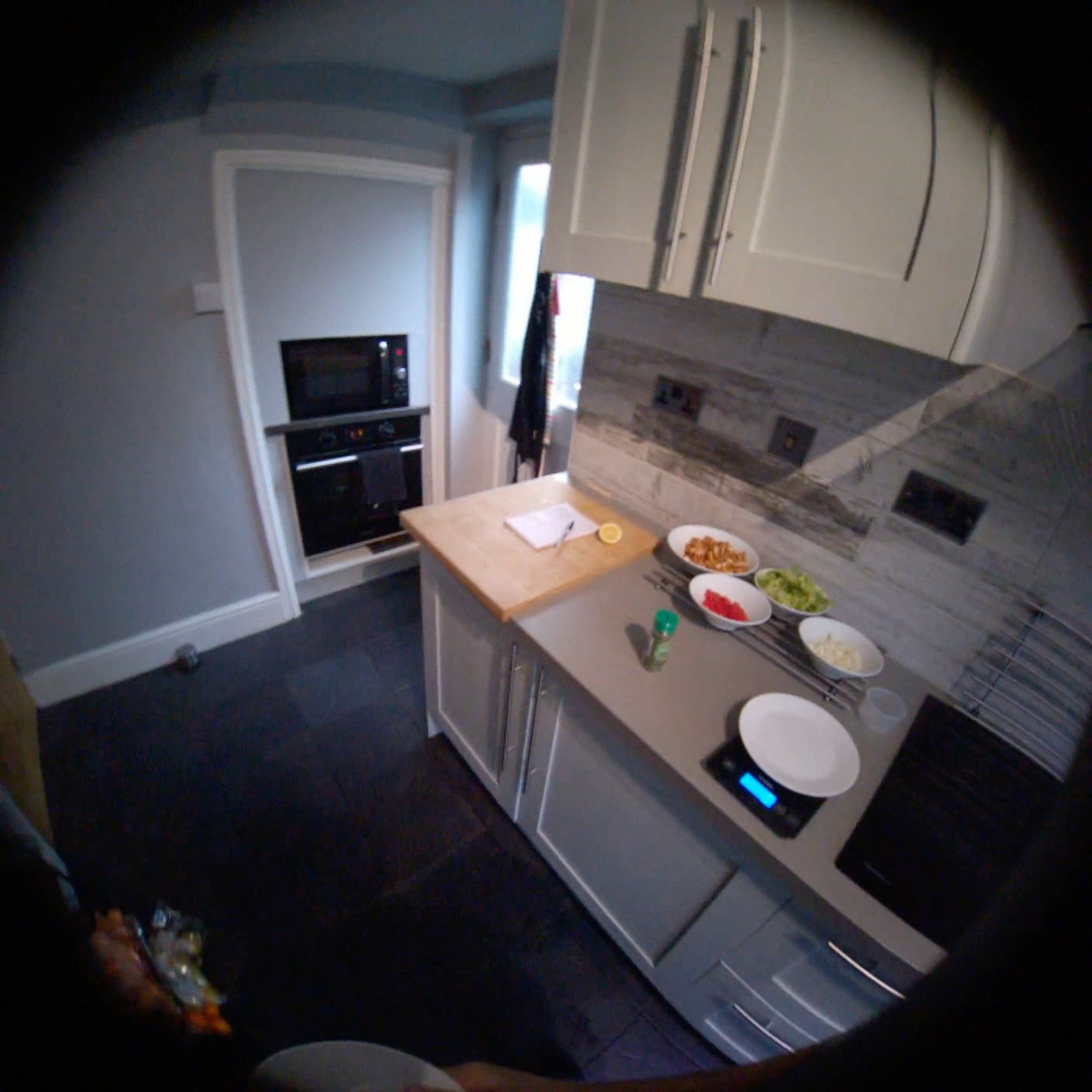}
            \colorbox{red!10}{\parbox{0.94\linewidth}{\centering\scriptsize\textcolor{red}{\textbf{The white serving bowl.}}}}
            \caption{}
        \end{subfigure}\hfill
        \begin{subfigure}[t]{0.48\linewidth}
            \centering
            \includegraphics[width=\textwidth]{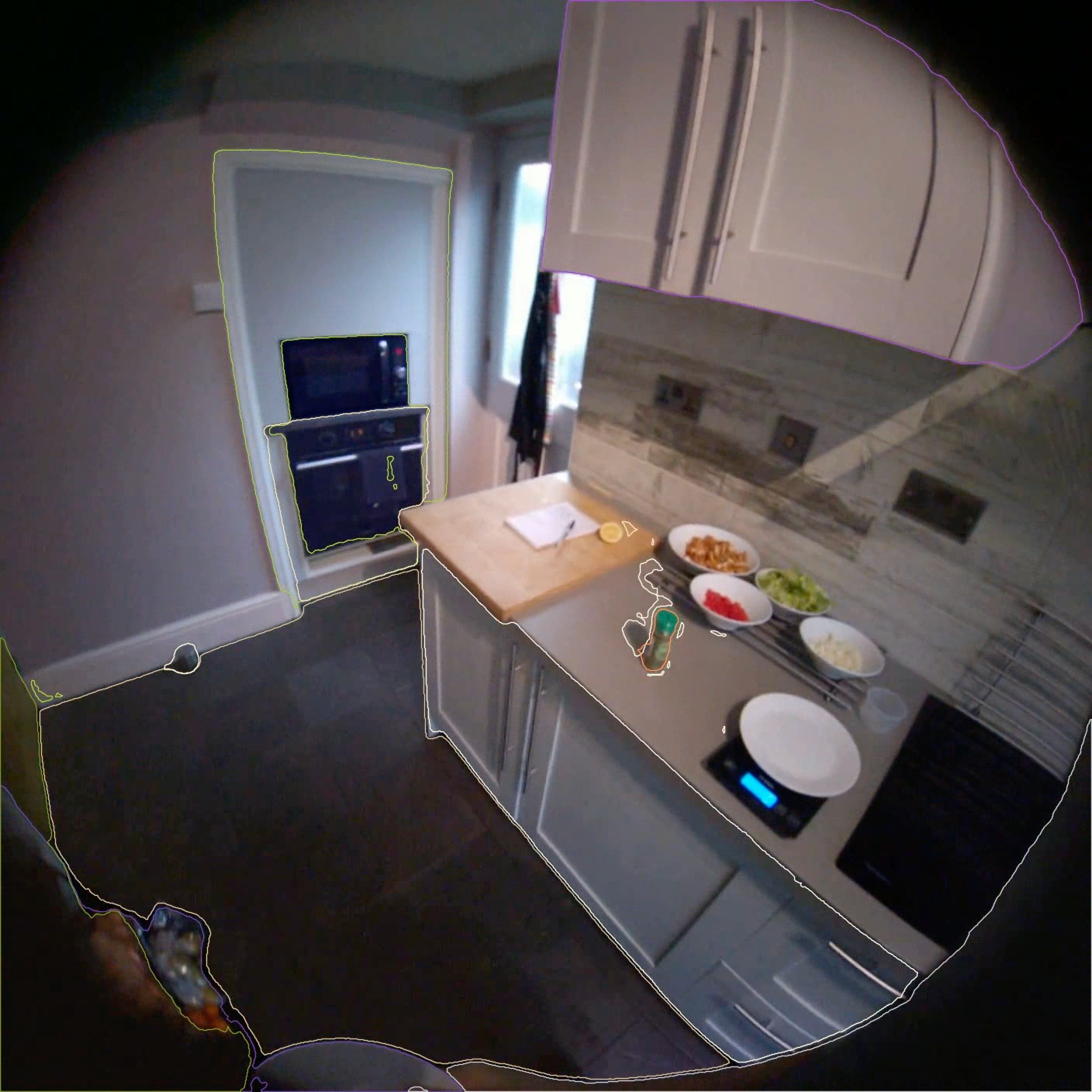}
            \colorbox{red!10}{\parbox{0.94\linewidth}{\centering\scriptsize\textcolor{red}{\textbf{The white serving bowl.}}}}
            \caption{}
        \end{subfigure}
                
        \begin{subfigure}[t]{0.48\linewidth}
            \centering
            \includegraphics[width=\textwidth]{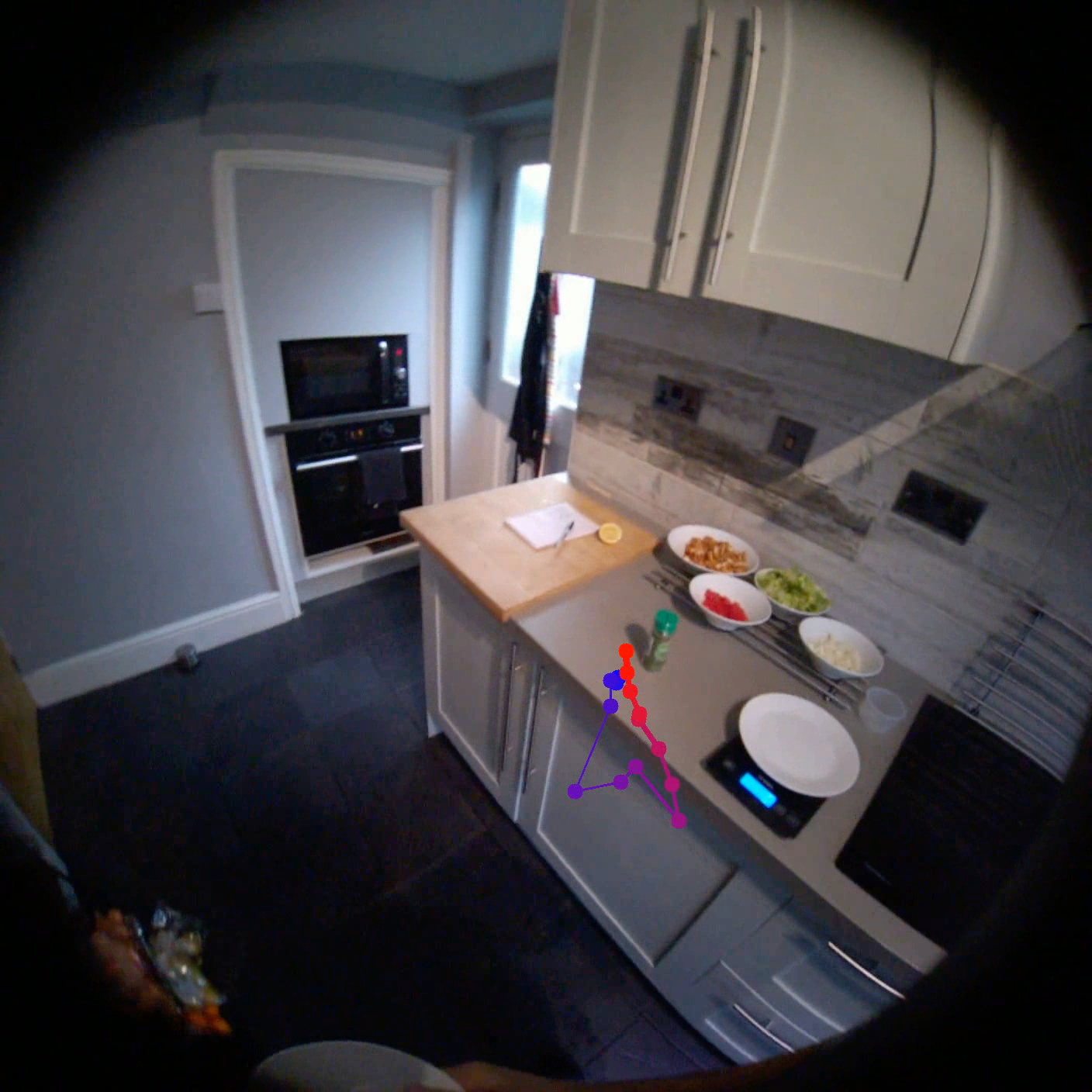}
            \colorbox{red!10}{\parbox{0.94\linewidth}{\centering\scriptsize\textcolor{red}{\textbf{The white serving bowl.}}}}
            \caption{}
        \end{subfigure}\hfill
        \begin{subfigure}[t]{0.48\linewidth}
            \centering
            \includegraphics[width=\textwidth]{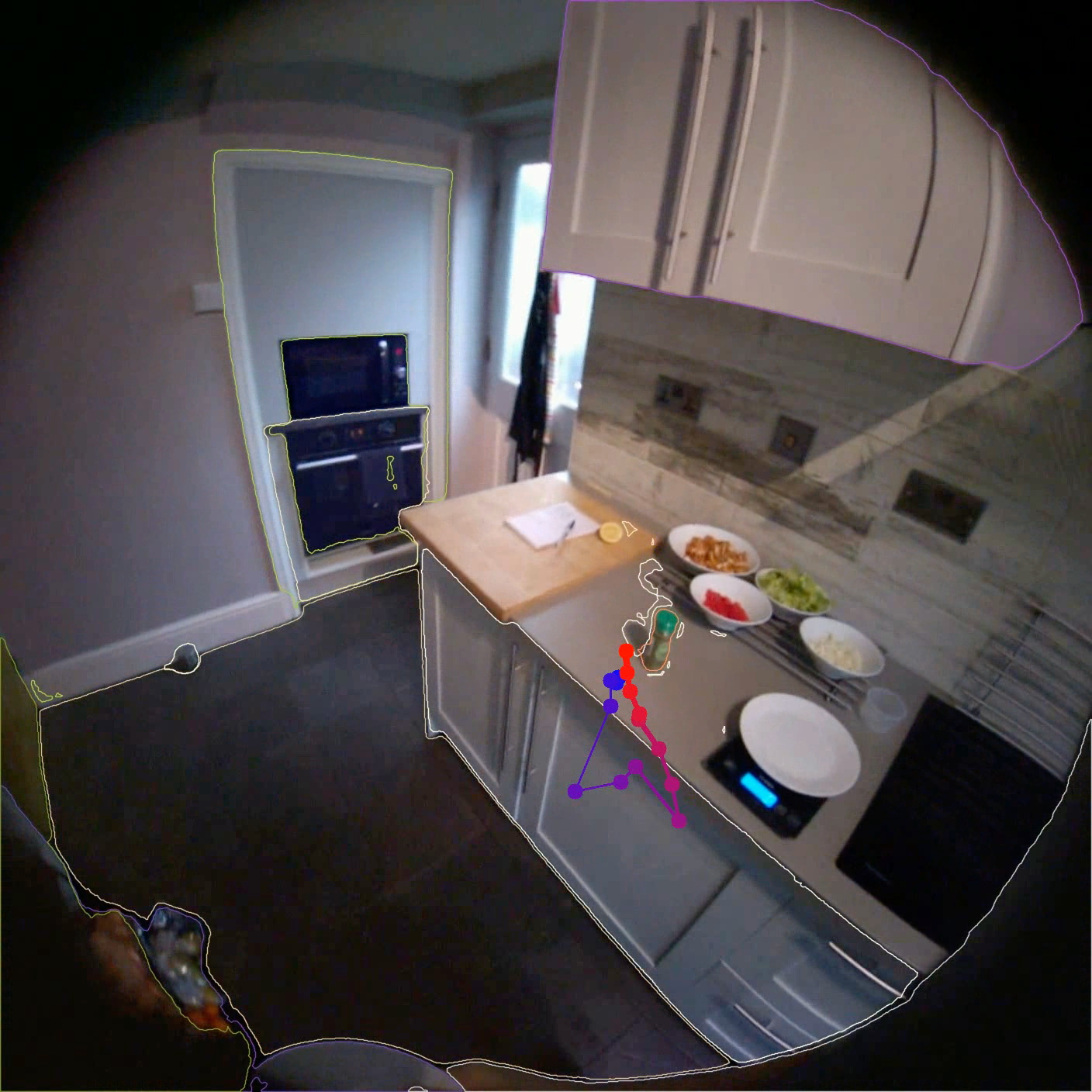}
            \colorbox{green!10}{\parbox{0.94\linewidth}{\centering\scriptsize\textcolor{green!60!black}{\textbf{The green capped condiment jar.}}}}
            \caption{}
        \end{subfigure}
    \end{minipage}
    
    \caption{A challenging scenario highlighting the positive effect of the synergy between \SoM and \textit{Gaze}.}
    \label{fig:qualitative_139}
\end{figure}

Figures~\ref{fig:qualitative_519}--\ref{fig:qualitative_139} provide qualitative examples showing the outputs of the different modules related to our architecture, highlighting their contribution to the final predictions.
Specifically, Figure~\ref{fig:qualitative_519} illustrates a case where, without our architecture, the VLLM fails to anticipate the correct human-object interaction (Figure~\ref{fig:qualitative_519}-a). Similarly, adding only the visual grounding capabilities through the SoM Module, is insufficient to anticipate the correct interaction (Figure~\ref{fig:qualitative_519}-b). Instead, incorporating gaze trajectory significantly enhances the model’s attention toward the correct object (Figure~\ref{fig:qualitative_519}-c and d).
Figure~\ref{fig:qualitative_139} illustrates the importance of jointly considering both SoM and Gaze Modules. In this example, neither SoM (Figure~\ref{fig:qualitative_139}-b) nor Gaze (Figure~\ref{fig:qualitative_139}-c) modules alone can disambiguate the target, resulting in an incorrect prediction. However, when both modules are exploited, the proposed method leads the VLLM to answer correctly (Figure~\ref{fig:qualitative_139}-d).

\subsection{Ablation Study}
\label{subsec:ablation}
We conducted a series of ablation studies to assess the contribution of each module within the proposed architecture, examining the impact of SoM on visual grounding, gaze on attention alignment, and different sampling strategies on accuracy. Since our architecture is independent of the adopted VLLM, we selected LLaVA-OneVision-7B for its computational efficiency and openness.

\begin{table}[t!]
\centering
  \caption{Impact of SoM, Gaze, and Sampling modules when adopting LLaVA-OneVision as the VLLM, with sample size $n = 15$ frames.}
  \begin{tabular}{lcccccc}
        \toprule
        \textbf{Strategy} & \textbf{$\lambda=0$} & \textbf{$\lambda=\tfrac{1}{100}$} & \textbf{$\lambda=\tfrac{1}{50}$} & \textbf{$\lambda=\tfrac{1}{25}$} & \textbf{$\lambda=\tfrac{1}{10}$} & \textbf{$\lambda=1$} \\
        \midrule
        VLLM only & 0.184 & 0.201 & 0.218 & 0.229 & 0.249 & 0.248 \\
        SoM & 0.188 & \textbf{0.223} & 0.234 & 0.231 & 0.248 & 0.258 \\
        Gaze & \textbf{0.203} & 0.210 & 0.220 & 0.230 & 0.252 & 0.256 \\
        SoM + Gaze & 0.201 & 0.222 & \textbf{0.243} & \textbf{0.249} & \textbf{0.272} & \textbf{0.261} \\
        \bottomrule
    \end{tabular}
\label{tab:ablation_modules_llava}
\end{table}

\subsubsection{Impact of individual modules.}
Table~\ref{tab:ablation_modules_llava} reports the impact of the different modules in our proposed architecture. Specifically, we compare the performance across varying values of the $\lambda$ parameter (columns) when adopting SoM, Gaze, or both modules (rows). As shown, predicting future human-object interactions using the \textit{VLLM only} strategy (first row) yields the worst results, regardless of the sampling strategy employed, obtaining an accuracy that ranges from 18.4\% (first column) to 24.9\% (fifth column). When $\lambda$ is equal to $0$ (first column) or $\frac{1}{100}$ (second column), which corresponds to a sampling strategy close to uniform, adopting only the Gaze Module or the SoM Module is sufficient, obtaining accuracies of 20.3\% (third row) and 22.3\% (second row), respectively. For all other cases, combining the SoM and Gaze modules (last row) yields the best performance. The optimal configuration is achieved with $\lambda = \frac{1}{10}$ (i.e., sampling frames close to the last frame of the observed past) when using both \textit{SoM + Gaze} modules, obtaining an accuracy of \textbf{27.2\%}.

\subsubsection{Sample size.}
Recent literature suggests that increasing the temporal size of the visual input in Vision LLMs does not necessarily lead to better performance~\cite{plizzari2025omniaegotempobenchmarkingtemporal}. Inspired by this observation, we conducted an ablation study showing that temporally oversampling the input video negatively impacts the accuracy of the proposed architecture. We performed experiments adopting the best configuration of our architecture which includes SoM and Gaze modules, varying the value of $\lambda$ of our Sampling Module. Table~\ref{tab:ablation_llava_sample_size} presents the results, indicating that the optimal configuration is achieved with sample size $n = 15$ (third row) and $\lambda=\frac{1}{10}$ (fifth column) obtaining an accuracy of \textbf{27.2\%}. Notably, any different sample size $n$ leads to a performance drop, regardless of the value of $\lambda$.

\begin{table}[t!]
\centering
  \caption{Impact of the chosen sample size $n$ on LLaVA-OneVision’s performance. The strategy used here is \textit{SoM + Gaze}.}
  \begin{tabular}{ccccccc}
        \toprule
        \textbf{Sample size} & \textbf{$\lambda=0$} & \textbf{$\lambda=\tfrac{1}{100}$} & \textbf{$\lambda=\tfrac{1}{50}$} & \textbf{$\lambda=\tfrac{1}{25}$} & \textbf{$\lambda=\tfrac{1}{10}$} & \textbf{$\lambda=1$} \\
        \midrule
        $n=5$ & 0.196 & 0.225 & 0.228 & 0.240 & 0.250 & 0.235 \\
        $n=10$ & 0.214 & 0.228 & 0.250 & 0.254 & 0.259 & 0.240 \\
        $n=15$ & 0.201 & 0.222 & 0.243 & 0.249 & \textbf{0.272} & \underline{0.261} \\
        $n=20$ & 0.222 & 0.217 & 0.216 & 0.244 & 0.252 & 0.251 \\
	    $n=25$ & 0.212 & 0.215 & 0.222 & 0.235 & 0.241 & 0.251 \\
	    $n=30$ & 0.210 & 0.207 & 0.212 & 0.223 & 0.237 & 0.240 \\
        \bottomrule
    \end{tabular}
\label{tab:ablation_llava_sample_size}
\end{table}

\begin{table}[t]
\centering
\renewcommand{\arraystretch}{1.1}
\setlength{\tabcolsep}{8pt}
\small
\caption{Performance of Gemini 2.0 Flash across varying input fps.}
\begin{tabular}{lccccc}
\toprule
\textbf{Strategy} &
\textbf{1 fps} &
\textbf{2 fps} &
\textbf{4 fps} &
\textbf{8 fps} &
\textbf{16 fps} \\
\midrule
VLLM only & 0.205 & 0.254 & 0.256 & 0.226 & 0.218 \\
SoM & \textbf{0.207} & 0.220 & 0.242 & 0.202 & 0.181 \\
Gaze & 0.193 & 0.236 & 0.264 & 0.256 & 0.228 \\
SoM + Gaze & 0.153 & \textbf{0.275} & \textbf{0.266} & \textbf{0.273} & \textbf{0.271} \\
\bottomrule
\end{tabular}
\label{tab:ablation_fps_gemini}
\end{table}

\subsubsection{Gemini 2.0 Flash.}
Since Gemini’s API does not support custom frame sequences~\cite{google2025aiapi}, we are unable to apply our sampling strategy when using this VLLM in the proposed framework. In particular, the only parameter we can modify is the frame rate of the input video. Table~\ref{tab:ablation_fps_gemini} reports the performance of our method highlighting the impact of each module across different fps.
Results show that independently from the fps, using both SoM and Gaze modules, the proposed framework achieves the best performance, except when the frame rate is equal to 1 fps, when using only the SoM model is sufficient (second row). The best configuration for Gemini 2.0 Flash is achieved with $2$ fps when using both \textit{SoM + Gaze} modules, obtaining an accuracy of 27.5\%.

\section{Conclusion and Future Directions}
\label{sec:conclusion}
We presented a novel approach for anticipating human-object interactions in egocentric videos by leveraging Vision LLMs enhanced with spatial information, gaze-based human intent modeling and effective video sampling. Our ablation study confirms that combining visual markers with gaze cues outperforms single-modality configurations, and that the proposed \textit{inverse exponential} sampling strategy significantly enhances the model's ability to capture task-relevant temporal dynamics. Our method achieves state-of-the-art results on the \textit{HD-EPIC Gaze Interaction Anticipation} benchmark, outperforming both the original baseline and recent challenge winners. Future work could expand this approach to real-time scenarios and long-term action anticipation.

\section*{Acknowledgments}
This research is supported by Next Vision s.r.l., the project ``Future Artificial Intelligence Research" (FAIR) -- PNRR MUR Cod. PE0000013 -- CUP: E63C22001940006, and the Research Program PIAno di inCEntivi per la Ricerca di Ateneo 2024/2026, project ``Multi-Agent Simulator for Real-Time Decision-Making Strategies in Uncertain Egocentric Scenarios" -- University of Catania.

%\section*{Disclosure of Interests}
%The authors have no competing interests to declare that are
%relevant to the content of this article.

\bibliographystyle{splncs04}
\bibliography{references}

\end{document}